\documentstyle[aclap]{article}
\title{\vspace{-0.5in}Sense Tagging: \\Semantic Tagging with a Lexicon}
 
\author{Yorick Wilks and Mark Stevenson\\
Department of Computer Science,\\
University of Sheffield,\\
Regent Court, 211 Portobello Street,\\
Sheffield S1 4DP, UK\\
{\tt \{yorick, marks\}@dcs.shef.ac.uk}
}

\begin{document}
\maketitle
\vspace{-0.5in}
\begin{abstract}
Sense tagging, the automatic assignment of the appropriate
  sense from some lexicon to each of the words in a text, is a
  specialised instance of the general problem of semantic tagging by
  category or type. We discuss which recent word sense disambiguation
  algorithms are appropriate for sense tagging. It is our belief that
  sense tagging can be carried out effectively by combining several
  simple, independent, methods and we include the design of such a
  tagger. A prototype of this system has been implemented, correctly
  tagging 86\% of polysemous word tokens in a small test set,
  providing evidence that our hypothesis is correct.
\end{abstract}

\section{Sense tagging}

This workshop is about semantic tagging: marking each word
token\footnote{Often loosened to each \emph{content} word in a
  text.} in a text with some marker identifying its semantic category,
similar to the way a part-of-speech tagger assigns a grammatical
category to each token in a text. Our recent work has been concerned
with sense tagging, a particular instance of this problem. Sense
tagging is the process of assigning, to each content word in a text,
its particular sense from some lexicon. This differs from the more
general case of semantic tagging, where the tags for each word (type)
are not be specific to that type and do not correspond to word senses in a
lexicon. For example the tags may be broad semantic categories such as
{\tt HUMAN} or {\tt ANIMATE} or WordNet synsets.

Another, broader, class of algorithms are word sense disambiguation
(WSD) algorithms. By WSD algorithm we mean any procedure which carries
out semantic disambiguation on words, these may not necessarily be
tagging algorithms, in that they do not attempt to mark every token in
a text but may be restricted to disambiguating small sets of word
types.

Sense tagging is a difficult problem: each word (type) has its own set
of tags which may be quite large. This rules out approaches which rely
on a discriminator being created for each semantic tag which is then
applied to text, although this is a valuable technique when there are
a small number of tags which are broad semantic categories.

However, sense tagging is an extremely useful procedure to carry out
since the tags which are associated during sense tagging are
rich in knowledge and therefore likely to be extremely useful for
further processing. Indeed, the lack of reliable, large-scale, sense
taggers has often been blamed for the failure of machine translation
for the last 30 years.

In this paper we shall discuss some recent approaches to the WSD
problem and examine their usefulness for the more specialised task of
sense tagging. We then propose an approach which makes use of several
different types of information and report a partial implementation of
this system which produces very encouraging results.

\section{Recent Word Sense Disambiguation algorithms}

Recent word sense disambiguation (WSD) algorithms can be categorised
into two broad types:

\begin{enumerate} 
\item \label{app_1} WSD using information in an explicit lexicon.
  This is usually a Machine Readable Dictionary (MRD) such as the {\it
    Longman Dictionary of Contemporary English} (LDOCE) \cite{ldoce},
  WordNet \cite{Miller90} or hand-crafted. Recent examples of this
  work include \cite{Bruce92}, \cite{Bruce94}, \cite{McRoy92}.
  
\item WSD using information gained from training on some corpus. This
  approach can be further sub-classified:

\begin{enumerate}
\item \label{app_2a} Supervised training, where information is
  gathered from corpora which have already been semantically
  disambiguated. As such corpora are hard to obtain, usually requiring
  expensive hand-tagging, research in this area
  has concentrated on other forms of lexical ambiguities, eg.
  \cite{Gale92}.
  
\item \label{app_2b} Unsupervised training, where information is
  gathered from raw corpora which has not been semantically
  disambiguated. The best examples of this approach has been the
  resent work of Yarowsky -- \cite{Yar92}, \cite{Yar93}, \cite{Yar95}.
\end{enumerate} 
\end{enumerate}

These approaches are not mutually exclusive and there are, of course,
some hybrid cases, for example Luk \cite{Luk95} uses information in
MRD definitions (approach \ref{app_1}) and statistical information
from untagged corpora (approach \ref{app_2b}).

\section{Comparing Different Approaches}

Approach \ref{app_2a} is the least promising since text tagged with
word senses is practically non-existent and is both time consuming and
difficult to produce manually. Much of the research in this area has
been compromised by the fact that researchers have focussed on
lexical ambiguities that are not true word sense distinctions, such
as words translated differently across two languages \cite{Gale92} or
homophones\footnote{Words pronounced identically but spelled
  differently.} \cite{Yar93}.

Even in the cases where data with the appropriate sense distinctions
is available, the text is unlikely to be from the desired domain: a
word sense discriminator trained on company news text will be much
less effective on text about electronics products. A discriminator
trained on many types of text so as to be generic will not be
particularly successful in any specific domain.

Approach \ref{app_2b} has received much attention recently. Its
disadvantage is that sense disambiguation is not carried out relative
to any well defined set of senses, but rather an ad hoc set. Although
this research has been the most successful of all approaches, it is
difficult to see what use could be made of the word sense distinctions
produced.

Using approach \ref{app_1} with hand crafted lexicons has the
disadvantage of being expensive to create: indeed Small and Rieger
\cite{Small82} attempted WSD using ``word experts'', which were
essentially hand crafted disambiguators. They reported that the word
expert for ``throw'' is ``currently six pages long, but should be ten
times that size'', making this approach impractical for any system
aiming for broad coverage.

\section{Proposed Approach}

Word senses are not absolute or Platonic but defined by a given
lexicon, as has been known for many years from early work on WSD, even
though the contrary seems widely believed: ``.. it is very difficult to
assign word occurrences to sense classes in any manner that is both
general and determinate. In the sentences ``I have a stake in this
country.'' and ``My stake in the last race was a pound'' is ``stake''
being used in the same sense or not? If ``stake'' can be interpreted
to mean something as vague as 'Stake as any kind of investment in any
enterprise' then the answer is yes. So, if a semantic dictionary
contained only two senses for ``stake'': that vague sense together
with 'Stake as a post', then one would expect to assign the vague
sense for both the sentences above. But if, on the other hand, the
dictionary distinguished 'Stake as an investment' and 'Stake as an
initial payment in a game or race' then the answer would be expected
to be different. So, then, word sense disambiguation is relative to
the dictionary of sense choices available and can have no absolute
quality about it.'' \cite{Wilks72}

There is no general agreement over the number of senses appropriate
for lexical entries: at one end of the spectrum Wierzbicka
\cite{Wier89} claims words have essentially one sense while
Pustejovsky believes that "... words can assume a potentially infinite
number of senses in context."\cite{Pust95} How, then, are we to get an
initial lexicon of word senses? We believe the best resource is still
a Machine Readable Dictionary: they have a relatively
well-defined set of sense tags for each word and lexical coverage is
high.

MRDs are, of course, normally generic, and much practical WSD work is
for sub-domains. We are adhering to the view that it is better to
start with such a generic lexicon and adapt it automatically with
specialist words and senses. The work described here is part of ECRAN
\cite{ecran}, a European LRE project on tuning lexicons to domains,
with a general sense tagging module used as a first stage.

\section{Knowledge Sources}

An interesting fact about recent word sense disambiguation algorithms
is that they have made use of different, orthogonal, sources of
information: the information provided by each source seems independent
of and has no bearing on any of the others. We propose a tagger that
makes use of several types of information (dictionary definitions,
parts-of-speech, domain codes, selectional preferences and collocates)
in the tradition of McRoy \cite{McRoy92} although, the information
sources we use are orthogonal, unlike the sources she used,
making it easier to evaluate the performance of the various modules.

\subsection{Part-of-speech}\label{pos}

It has already been shown that part-of-speech tags are a useful
discriminator for semantic disambiguation \cite{Wilks96b}, although
they are not, normally, enough to fully disambiguate a text.  For
example knowing "bank" in "My bank is on the corner." is being used as
a noun will tell us that the word is not being used in the 'plane
turning corner' sense but not whether it is being used in the
'financial institution' or 'edge of river' senses. Part-of-speech tags
can provide a valuable step towards the solution to sense tagging:
fully disambiguating about 87\% of ambiguous word tokens and reducing
the ambiguity for some of the rest.

\subsection{Domain codes (Thesaural categories)}\label{domain_codes}

Pragmatic domain codes can be used to disambiguate (usually nominal)
senses, as was shown by \cite{Bruce92} and \cite{Yar92}. Our intuition
here is that disambiguation evidence can be gained by choosing senses
which are closest in a thesaural hierarchy. Closeness in such a
hierarchy can be effectively expressed as the number of nodes between
concepts. We are implementing a simple algorithm which prefers close
senses in our domain hierarchy which was derived from LDOCE \cite{Bruce92}.

\subsection{Collocates}

Recent work has been done using collocations as semantic
disambiguators, \cite{Yar93}, \cite{Dorr96}, particularly for verbs.
We are attempting to derive disambiguation information by examining
the prepositions as given in the subcategorization frames of verbs,
and in the example sentences in LDOCE.

\subsection{Selectional Preferences}
 
There has been a long tradition in NLP of using selectional
preferences for WSD \cite{Wilks72}. This approach has been
recently used by \cite{McRoy92} and \cite{Mahesh96}. At its best it
disambiguates both verbs, adjectives and the nouns they modify at the
same time, but we shall use this information late in the
disambiguation process when we hope to be reasonably confident of the
senses of nouns in the text from processes such as \ref{domain_codes}
and \ref{dict_def}.

\subsection{Dictionary definitions}\label{dict_def}

Lesk \cite{Lesk86} proposed a method for semantic disambiguation using
the dictionary definitions of words as a measure of their semantic
closeness and proposed the disambiguation of sentences by computing
the overlap of definitions for a sentence. Simmulated annealing, a
numerical optimisation algorithm, was used to make this process
practical \cite{Cowie92}, choosing an assignment of senses from as many as
$10^{10}$ choices. 

The optimisation is carried out by minimising an evaluation function,
computed from the {\em overlap} of a given configuration of senses.
The overlap is the total number of times each word appears more than
once in the dictionary definitions of all the senses in the
configuration. So that if the word ``bank'' appeared three times in a
given configuration we would add two to the overlap total. This
function has the disadvantage that longer definitions are prefered
over short ones, since these simply have more words which can
contribute to the overlap. Thus short definitions or definitions by
synonym are penalised.

We attempted to solve this problem by making a slight change to the
method for calculating the overlap. Instead of each word contributing
one we normalise it's contribution by the number of words in the
definition it came from, so if a word came from a definition with three
words it would add one third to the overlap total. In this way long
definitions have to have many words contributing to the total to be
influential and short definitions are not penalised.

We found that this new function lead to a small improvement in the
results of the disambiguation, however we do not believe this to be
statistically significant.

\section{A Basic Tagger}
        
We have recently implemented a basic version of this tagger, initially
incorporating only the part-of-speech (\ref{pos}) and dictionary
definition (\ref{dict_def}) stages in the process, with further stages
to be added later. Our tagger currently consists of three modules:

\begin{itemize}
\item Dictionary look-up module
\item Part-of-speech filter
\item Simulated annealing 
\end{itemize}

\begin{enumerate}
\item We have chosen to use the machine readable version of LDOCE as
  our lexicon. This has been used extensively in NLP research and
  provides a broad set of senses for sense tagging.
  
  The text is initially stemmed, leaving only morphological roots, and
  split into sentences. Then words belonging to a list of stop words
  (prepositions, pronouns etc.) are removed. For each of the remaining
  words, each of its senses are extracted from LDOCE and stored with
  that word. The textual definitions in each sense is processed to
  remove stop words and stem remaining words.
  
\item The text is tagged using the Brill tagger \cite{Brill92} and a
  translation is carried out using a manually defined mapping from the
  syntactic tags assigned by Brill (Penn Tree Bank tags \cite{ptb})
  onto the simpler part-of-speech categories associated with LDOCE
  senses. We then remove all senses whose part-of-speech is not
  consistent with the one assigned by the tagger, if none of the
  senses are consistent with the part-of-speech we assume the tagger
  has made an error and do not remove any senses.

\item The final stage is to use the simulated annealing algorithm to
  optimise the dictionary definition overlap for the remaining senses.
  This algorithm assigns a single sense to each token which is the tag
  associated with that token.
\end{enumerate}

\section{Example Output}

Below is an example of the senses assigned by the system for the
sentence ``A rapid rise in prices soon eventuated unemployment.'' We
show the homograph and sense numbers from LDOCE with the stemmed
content words from the dictionary definitions which are used to
calculate the overlap following the dash.

\begin{itemize}
\item {\tt rapid} homograph 1 sense 2 -- done short time
\item {\tt rise} homograph 2 sense 1 -- act grow greater powerful
\item {\tt soon} homograph 0 sense 1 -- long short time
\item {\tt prices} homograph 1 sense 1 -- amount money which thing be
  offer sell buy
\item {\tt unemployment} homograph 0 sense 1 -- condition lack job 
\end{itemize}

The senses have additional information associated which we do not show
here: domain codes, part of speech and grammatical information as well
as semantic information.

The senses for a word in LDOCE are grouped into homographs, sets of
senses realeated by meaning. For example, one of the homographs of
``bank'' means roughly 'things piled up', the different senses
distinguishing exactly what is piled up.

\section{Results}

We have conducted some preliminary testing of this approach: our tests
were run on 10 hand-disambiguated sentences from the Wall Street
Journal amounting to a 209 word corpus. We found that of, the word
tokens which had more than 1 homograph, 86\% were assigned the correct
homograph and 57\% of tokens were assigned the correct sense using our
simple tagger. These figures should be compared to 72\% correct
homograph assignment and 47\% correct sense assignment using simulated
annealing alone on the same test set (see \cite{Cowie92}). It should
be noted that the granularity of sense distinctions at the LDOCE
homograph level (eg. ``bank'' as 'edge of river' or 'financial
institution') is the same as the distinctions made by current
small-scale WSD algorithms (eg.  \cite{Gale92}, \cite{Yar93},
\cite{Schutze92}) and our system is a true tagging algorithm,
operating on free text.

Our evaluation is unsatisfactory due to the small test set, but does
demonstrate that the use of independent knowledge sources leads to an
improvement in the quality of disambiguation. We fully expect our
results to improve with the addition of further, independent, modules.

\section{Conclusion}

In this paper we have argued that semantic tagging can be carried out
only relative to the senses in some lexicon and that a machine readable
dictionary provides an appropriate set of senses.

We reported a simple semantic tagger which achieves 86\% correct
disambiguation using two independent sources of information:
part-of-speech tags and dictionary definition overlap. A proposal to
extend this tagger is developed, based on other, mutually independent,
sources of lexical information.

\section*{Acknowledgements}

This research was supported by the European Union Language Engineering
project ECRAN, number LE2110. We are grateful to Jim Cowie at CRL for
providing the simulated annealing code and to Kevin Humphries and
Hamish Cunningham of the Sheffield NLP group for advice with the
implementation of the tagger.

\end{document}